\documentclass[10pt,twocolumn,letterpaper]{article}

\usepackage{cvpr}              

\usepackage{graphicx}
\usepackage{amsmath}
\usepackage{amssymb}
\usepackage{multirow}
\usepackage{booktabs}
\usepackage{algorithm}
\usepackage{algorithmic}
\usepackage[accsupp]{axessibility}

\DeclareMathOperator*{\argmax}{argmax}

\usepackage[pagebackref,breaklinks,colorlinks]{hyperref}

\usepackage[capitalize]{cleveref}
\crefname{section}{Sec.}{Secs.}
\Crefname{section}{Section}{Sections}
\Crefname{table}{Table}{Tables}
\crefname{table}{Table}{Tabs.}

\newcommand{\comment}[1]{}

\def\cvprPaperID{4908} 
\def\confName{CVPR}
\def\confYear{2022}

\begin{document}
\definecolor{commentcolor}{RGB}{110,154,155}   
\newcommand{\PyComment}[1]{\ttfamily\textcolor{commentcolor}{\# #1}}  
\newcommand{\PyCode}[1]{\ttfamily\textcolor{black}{#1}} 
\title{Leverage Your Local  and  Global Representations: A New Self-Supervised Learning Strategy}

\author{
Tong Zhang$^{1}$ ~
Congpei Qiu$^{2}$ ~
Wei Ke$^{2}$ ~
Sabine Süsstrunk$^{1}$ ~
Mathieu Salzmann$^{1}$\\

$^1$ School of Computer and Communication Sciences, EPFL, Switzerland \\
$^2$ Xi'an Jiaotong University, China\\
}
\maketitle


\begin{abstract}
Self-supervised learning (SSL) methods aim to learn view-invariant representations by maximizing the similarity between the features extracted from different crops of the same image regardless of cropping size and content. In essence, this strategy ignores the fact that two crops may truly contain different image information, e.g., background and small objects, and thus tends to restrain the diversity of the learned representations.
In this work, we address this issue by introducing a new self-supervised learning strategy, LoGo, that explicitly reasons about {\bf Lo}cal and {\bf G}l{\bf o}bal crops. To achieve view invariance, LoGo encourages similarity between global crops from the same image, as well as between a global and a local crop. However, to correctly encode the fact that the content of smaller crops may differ entirely, LoGo promotes two local crops to have dissimilar representations, while being close to global crops. Our LoGo strategy can easily be applied to existing SSL methods. Our extensive experiments on a variety of datasets and using different self-supervised learning frameworks validate its superiority over existing approaches. Noticeably, we achieve better results than supervised models on transfer learning when using only $1/10$ of the data.\footnote{Our code and pretrained models can be found at https://github.com/ztt1024/LoGo-SSL.
Correspondence to Ke Wei (wei.ke$@$mail.xjtu.edu.cn).}


\end{abstract}



\section{Introduction}


\begin{figure*}
  \centering
  \begin{subfigure}[b]{0.57\linewidth}
        \includegraphics[width=1.0\linewidth]{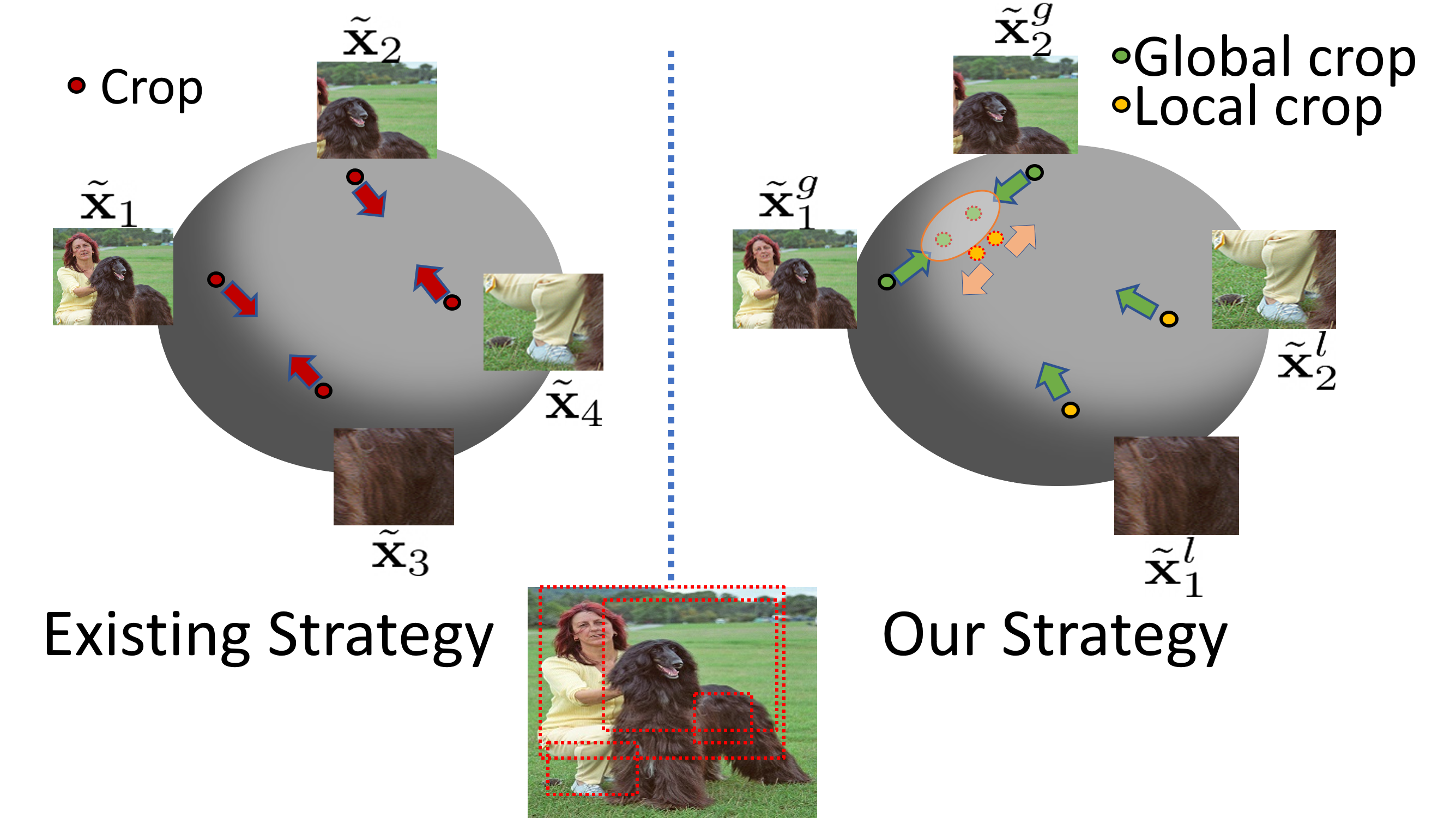}
    \caption{}
    \label{fig:overview-a}
  \end{subfigure}
  \hfill
  \begin{subfigure}[b]{0.42\linewidth}
        \includegraphics[width=1\linewidth]{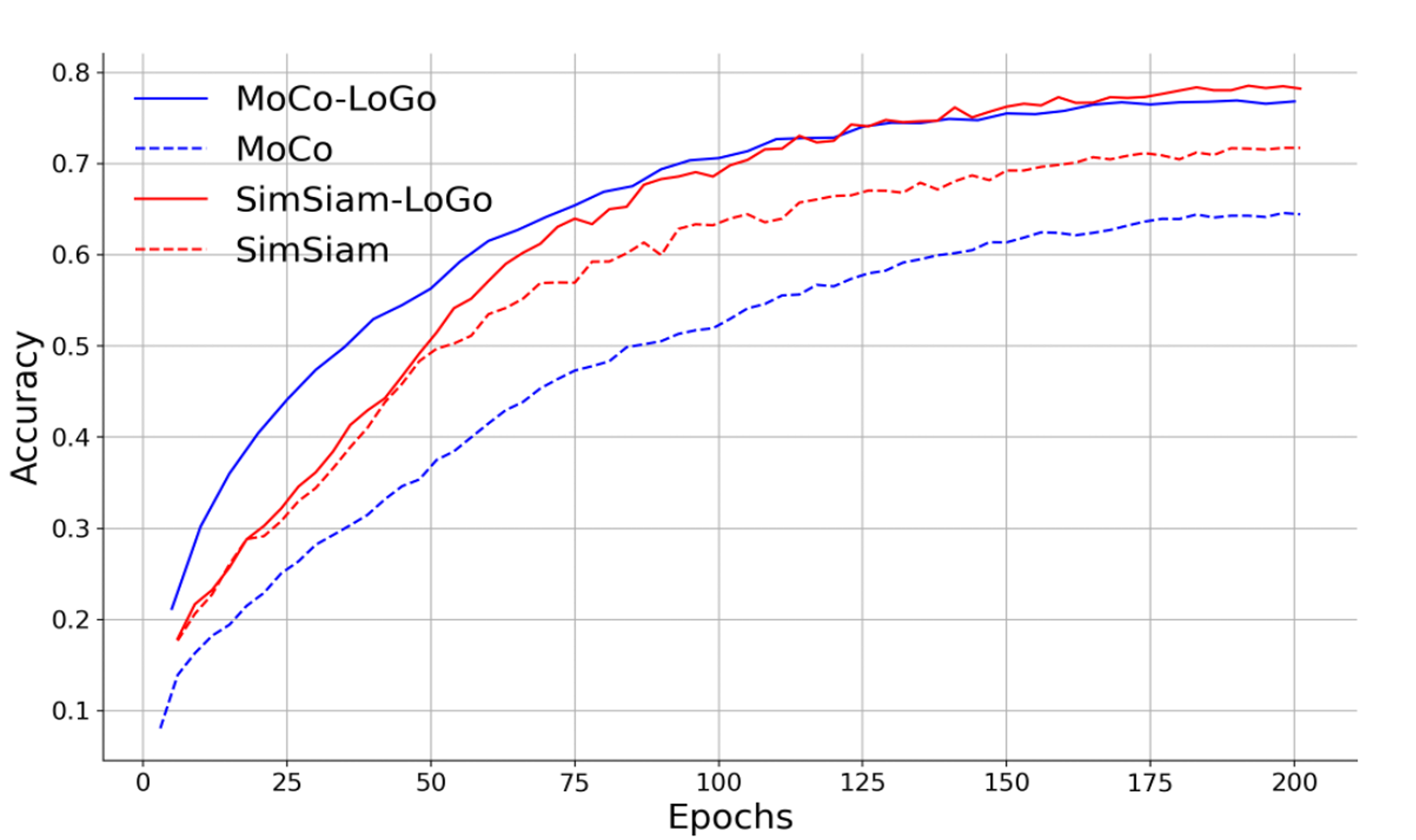}
    \caption{}
    \label{fig:overview-b}
  \end{subfigure}
  \caption{(a) Overview of our self-supervised learning strategy. To learn a view-invariant representation that nonetheless encodes semantic information about local objects, we seek to maximize the similarity between global crops while allowing the local crops to remain distant from each other, thus accounting for the fact that local crops may represent entirely different objects. (b) Monitoring of the KNN top-1 accuracy on ImageNet-100 with a ResNet-34 backbone evidences the benefits of our approach in different SSL strategies.
  }
  \label{fig:overview}
\end{figure*}




 Building on the great success of supervised learning in visual tasks such as image classification~\cite{lecun1989backpropagation, krizhevsky2012imagenet,he2016deep} and object detection~\cite{he2017mask,girshick2015fast}, significant efforts have recently been dedicated to learning high-level representations without human annotations. Inspired by the pre-training stage in natural language processing, e.g.\ GPT~\cite{radford2018improving} and BERT~\cite{devlin2018bert}, such a self-supervised learning (SSL) approach aims to learn representations that extract useful information for a downstream task in an unsupervised manner, thus providing an effective initialization to start from when some annotated data for the downstream tasks become available. Recently, SSL has been proven to be as effective as supervised pre-training, or even more effective in some cases~\cite{chen2020improved,caron2020unsupervised}.
 
The basic principle behind existing SSL approaches can be traced back to~\cite{linsker1988self, hadsell2006dimensionality} and consists of learning a representation that is shared across different views of the same input, yet carries discriminative information. In vision tasks, this is typically achieved by maximizing the similarity between two augmented views of the same image while penalizing trivial solutions using various techniques. For example, contrastive learning~\cite{chen2020simple, he2020momentum} incorporates negative pairs, where one view comes from a different image, to prevent the network from constantly generating the same output; non-contrastive methods~\cite{grill2020bootstrap,chen2021exploring} only rely on positive pairs by modifying the back-propagation mechanism to prevent collapse; clustering-based methods~\cite{asano2019self,caron2020unsupervised} perform online clustering to keep the consistency between exemplar representations (the centroids of clusters) and different views of the same image.

Intuitively, one should expect the representations of random crops with smaller sizes to have a larger variance than that of larger crops because, as shown in Figure~\ref{fig:overview}, they may truly encode entirely different content. Nevertheless, existing methods encourage \emph{all} the random crops of the same image to have similar representations. This complicates the learning process and tends to lead the network to discarding valuable image information to achieve such invariance. This was, for example, observed in~\cite{caron2021emerging}, where the multi-crop strategy of ~\cite{caron2020unsupervised} was shown to yield a performance drop when applied to other SSL methods, such as BYOL~\cite{grill2020bootstrap}, SimSiam~\cite{chen2021exploring}, and MoCo~\cite{he2020momentum}

In this paper, we address this limitation by introducing a  new multi-crop SSL strategy, LoGo,  which exploits the relationships between {\bf lo}cal and {\bf g}l{\bf o}bal image patches in different, well-adapted ways, and can easily be integrated into existing SSL frameworks. Specifically, we exploit two different kinds of crops: Large ones that encompass a global view of the input image, thus being well-suited to learn a view-invariant representation; and small ones with a higher variance that focus on local image regions, thus allowing the model to encode information such as background, texture, and objects. As illustrated in Figure~\ref{fig:overview}, we then design a loss function that (i) pulls the global representations of the same image close to each other, while also encouraging each local representation of that image to be close to the global ones; (ii) favors the different local representations to remain distant to account for the differences between the local patches. Altogether, this provides the model with the flexibility to keep apart the local representations that encode different regions while nonetheless encouraging the representations of all crops from the same image to cluster in the latent space.

Furthermore, to account for the fact that traditional distance metrics may be unreliable in a high dimensional space~\cite{aggarwal2001surprising}, we introduce a new approach to evaluate the similarity between the representations of two patches. Specifically, based on the assumption that the similarity of two local crops from the same image is greater than that of two local crops from different images with high probability, we train an MLP to discriminate between pairs of local crops from the same or from different images, and exploit its prediction as a similarity score. 

Our contributions can be summarized as follows:
 \begin{itemize}
 
    \item We exploit both global and local views in SSL to encode rich semantic information. To this end, we encourage similarity across global crops to achieve view invariance, but allow the local crops to be dissimilar to maintain the diversity of local object representations. 
    
    \item We introduce a learnable similarity measure to overcome the limitations of standard metrics in high dimensional feature space.
    
    \item Our approach generalizes to different SSL frameworks, including constrastive (e.g., MoCo~\cite{he2020momentum}) and non-contrastive (e.g., SimSiam~\cite{chen2021exploring}) ones.
    
    \item Our approach  allows the network to be trained on smaller datasets, which benefits downstream tasks where the training-testing domain gap is large.

\end{itemize}
We demonstrate the benefits of our approach over the state-of-the-art SSL techniques on several datasets. Importantly, our strategy enables the self-supervised models to surpass their supervised counterparts on dense prediction tasks with only 1/10 of the training data.

\section{Related Work}
SSL or representation learning frameworks can be roughly grouped into two categories: Those that are trained on pretext tasks, such as solving jigsaw puzzles~\cite{noroozi2016unsupervised} or predicting color from grayscale images~\cite{zhang2016colorful}, 
and those that optimize different learning objectives.
Our work falls in this second category, and we, therefore, focus the discussion below on the methods that also do.

\textbf{Contrastive learning methods.} 
 Contrastive learning aims to maximize a notion of affinity between pairs of positive samples while minimizing the affinity between negative pairs. This is typically achieved by optimizing the InfoNCE loss~\cite{oord2018representation}. To obtain diverse and discriminative feature representations, contrastive learning typically leverages data augmentation. For example,  Deep InfoMax~\cite{hjelm2018learning}, and its multi-scale version~\cite{bachman2019learning} aim to maximize the mutual information between the global and local features of an input image, that is, the feature vectors of the last layer after global pooling and the ones across all the channels in each location.
 Their positive pairs are defined using a single view of an image, which limits the diversity of the learned representation. CMC~\cite{tian2020contrastive} maximizes the mutual information between the feature representations of different modalities, e.g. semantic map, YCbCr, or depth map of the same image.
SimCLR~\cite{chen2020simple} is the first to augment each image twice and create positive pairs of distorted-original images and negative pairs using two different images. MoCo~\cite{he2020momentum} improves the contrastive training by using a memory bank to store negative pairs and avoid degenerate solutions. MoCo-V2~\cite{chen2020improved} shows that stronger augmentations and the use of multiple crops boost the performance of self-supervised learning. Furthermore, Wang \& Isola~\cite{wang2020understanding} provide theoretical proof of reinterpreting the InfoNCE loss as two terms: aligning features that belong to the same instance and spreading normalized learned features on a hypersphere. However, the theory can only be applied to the contrastive case and the empirical performance improvement is marginal.

\textbf{Non-contrastive learning methods.} 
One of the main difficulties in contrastive learning consists of defining meaningful negative pairs. To counteract this, BYOL~\cite{grill2020bootstrap} demonstrates that using only positive pairs is sufficient to avoid degenerate solutions when exploiting a Siamese network with one branch acting as a momentum encoder and used to supervise the other branch. Subsequently, SimSiam~\cite{chen2021exploring} proposes a simpler Siamese network, arguing that momentum is not required but that a predictor and stop-gradient are. This approach appends a predictor to one branch of the Siamese backbone and stops the gradient of that branch from being back-propagated to the backbone.


\textbf{Clustering-based methods.} Clustering itself has been an important research direction in unsupervised learning~\cite{zhang2019neural,ji2017deep,xie2016unsupervised,yang2016joint,caron2018deep,chang2017deep,zhang2018scalable}, and is nowadays used for representation learning. For example, DeepCluster~\cite{caron2018deep} alternately clusters the learned representations and predicts the cluster assignments; SeLa~\cite{asano2019self} simultaneously learns the representation and the cluster assignments by using the Sinkhorn-Knopp algorithm to perform online updates; SwAV~\cite{caron2020unsupervised} utilizes the same technique within a Siamese network to compute soft assignments from one view, which supervise the 
feature distribution in the other view. SwAV~\cite{caron2020unsupervised} further demonstrates that using multiple crops for each image helps their training. However, SwAV does not reason about the potential lack of shared information between multiple local crops, which is what we achieve here. Furthermore, the above-mentioned methods require either additional memory bank~\cite{caron2018deep,asano2019self} or very large batch sizes~\cite{caron2020unsupervised} to yield a stable and robust optimization.

Recently, the transformer-based Dino~\cite{caron2021emerging} network, a follow-up work of SwAV, proposes to use global views as teachers to supervise the local views' probability-like representation. 
However, this method inherently encourages the local crops to have similar representations to the global ones even though they may contain different objects.

In short, all of the existing methods encourage all the crops, regardless of their actual semantic information, to have similar representations. As such, to achieve view invariance, they tend to discard relevant semantic information, thus undermining the ability to transfer the resulting representations to downstream tasks. Here, we, therefore, propose a new SSL strategy that addresses this limitation.



\section{Methodology}
Our goal is to develop a self-supervised learning approach that is able to handle complex images depicting objects of different semantics. We aim for our approach to be general, and thus applicable to both contrastive and non-contrastive learning strategies. Therefore, below, we first review the contrastive and non-contrastive paradigm together with a representative framework for each, namely MoCo~\cite{he2020momentum} and SimSiam~\cite{chen2021exploring}. Subsequently, we introduce our hierarchical local-global model and our approach to learning a similarity measure.

\vspace{0.1cm}
\textbf{Notation.}
We use $\tau^g$ and $\tau^l$ to denote the operation sets for global and local augmentation, with $r_g$ and $r_l$ denoting the lower bound of the global crops' size and the upper bound of the local crops' size, respectively. The global and local views, namely $\tilde{\mathbf{x}}^g$ and $\tilde{\mathbf{x}}^l$, are obtained by applying $\tau^g$ and $\tau^l$ to the same image  $\mathbf{x} \in \mathbb{R}^{W \times H}$, where $W$ and $H$ are the image width and height. Similarly, $\mathbf{z} \in \mathbb{R}^n$ denotes the latent representation obtained by the encoder function $f_{\theta_e}: \mathbb{R}^{W \times H} \rightarrow \mathbb{R}^{n}$, and $\mathbf{z}^+$ and $\mathbf{z}^-$ are its corresponding positive and negative counterpart, respectively.

\begin{figure*}
\centering
\includegraphics[scale=0.4]{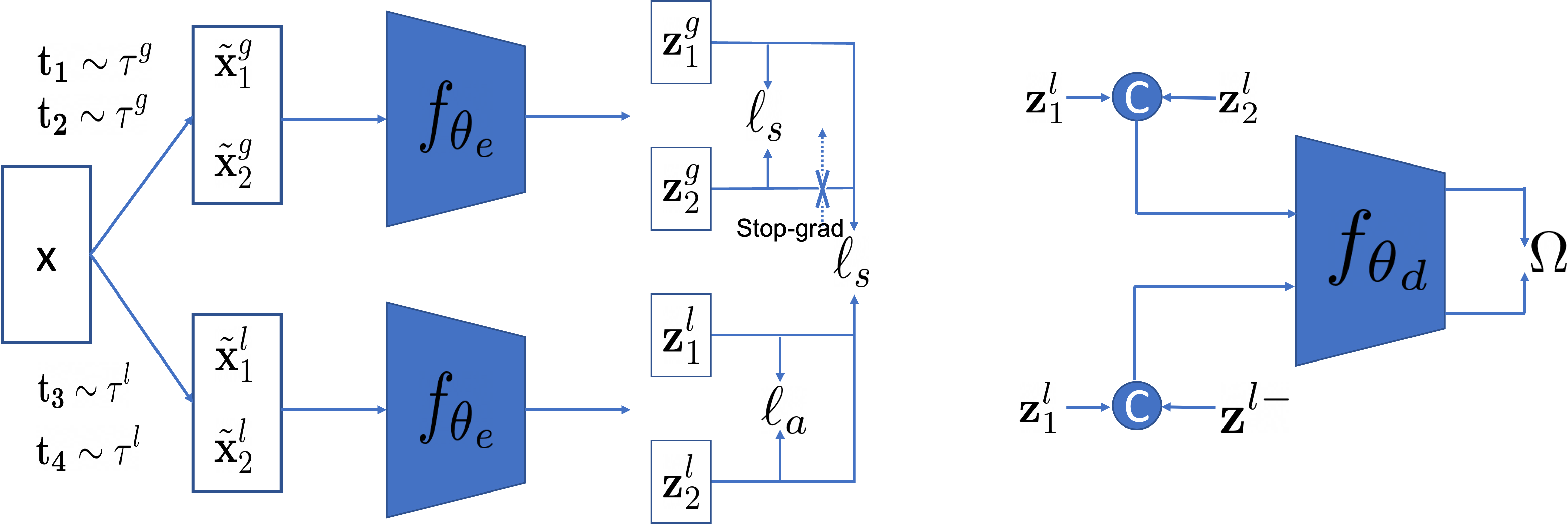}
\caption{Our LoGo structure (left) and local affinity measure $f_{\theta_d}$ (right).  $f_{\theta_e}$ represents the feature encoder, which includes a backbone network followed by a multi-layer perceptron. Each image is augmented into global and local crops which are fed to the encoder. We maximize the global-to-global and local-to-global similarity by optimizing $\ell_s$, which can be either the cosine or InfoNCE loss. Simultaneously, we maximize the dissimilarity between pairs of local crops by optimizing the output of a learned similarity measure $\ell_a$. Note that $\mathbf{z}^l$ is detached from the encoder, and no gradients are back-propagated to the encoder when training  $f_{\theta_d}$.}
\label{fig:framework}
\end{figure*}


\subsection{Similarity Loss}
Learning a feature representation without supervision is typically achieved by maximizing the similarity between the samples in positive pairs, while optionally minimizing the similarity of those in negative pairs. Our approach can be applied to most SSL techniques. To illustrate this, we therefore consider two typical similarity loss functions:  Info-NCE~\cite{chen2020simple,he2020momentum,oord2018representation,tian2020contrastive}, commonly-used in contrastive learning, and the cosine loss~\cite{chen2021exploring,grill2020bootstrap}, often employed in the non-contrastive scenario. 

Info-NCE was introduced by CPC~\cite{oord2018representation} and can be expressed as
\begin{equation}
\begin{split}
    \mathcal{L}^{NCE}(\mathbf{z},\mathbf{z}^+,\mathbf{z}^-) = - \log& \frac{\exp \left( \mathbf{z} \cdot \mathbf{z}^+ / \tau\right)}{ \exp \left( \mathbf{z} \cdot \mathbf{z}^+ / \tau \right) + \sum\exp \left( \mathbf{z} \cdot \mathbf{z}^- / \tau \right)}, \label{eqn:NCE}
\end{split}
\end{equation}
where $\tau$ is a temperature hyper-parameter, and $\mathbf{z}$ is the feature representation of augmented images encoded by $f_{\theta_e}$, i.e, $\mathbf{z} = f_{\theta_e}(\tilde{x})$. $\mathbf{z}^+$ is a positive sample and $\mathbf{z}^-$ is a negative one, which could be sampled from a memory bank~\cite{he2020momentum} or obtained using a large batch size~\cite{chen2020simple}.

By contrast, the cosine loss used in SSL does not exploit negative samples. It can be written as
\begin{equation}
    \mathcal{L}^{cos}(\mathbf{z}_{1},\mathbf{z}_{2}) =  -\frac{h(\mathbf{z}_1)}{\left\| h(\mathbf{z}_1)\right\|_{2}} \cdot \frac{\mathbf{z}_{2}}{\left\|\mathbf{z}_{2}\right\|},\label{eqn:cos}
\end{equation}
where $h$ is an MLP layer used to predict the ``mean" of the set of positive samples for $\mathbf{z}$. In this context, SimSiam~\cite{chen2021exploring} uses Siamese networks and stops the back-propagation for the $\mathbf{z}_{2}$ branch, whereas BYOL~\cite{grill2020bootstrap} uses a momentum encoder to update the encoder parameters.

\subsection{Our Approach}
In the presence of complex image content, such as multiple objects, existing approaches to generating positive, and optionally negative pairs, suffer from several drawbacks. First, depending on the random cropping, two different views of the same image might depict entirely different content. Conversely, two different images may share some content, and thus crops from these different images might in fact depict the same object category. Directly applying existing SSL strategies yields highly noisy and potentially contradictory constraints, thus complicating the learning process.

To address this, we exploit two different kinds of crops, local and global ones. Specifically, for each input image $\mathbf{x}$, we extract to two global views $\tilde{\mathbf{x}}^g_{1,2}$ and two local views $\tilde{\mathbf{x}}^l_{1,2}$ from the augmentation sets $\tau^g$ and $\tau^l$, respectively. We then optimize the global-to-global, local-to-global, and local-to-local relationships respectively. Note that, below, we use $\ell_s$ to denote a general similarity loss, which in our experiments will be either Eq.~\ref{eqn:NCE} or Eq.~\ref{eqn:cos}.



\textbf{Global-to-global.} Because the global views encompass most of the semantic content of the original image, we aim to reach a consensus among their representations by maximizing the similarity between global views from the same image, while optionally minimizing the similarity between the global views of different images in contrastive cases.
We therefore write a global-to-global loss as
\begin{equation}\label{eqn:g2g}
    \mathcal{L}_{gg} = \mathbb{E}_{\mathbb{P}_{\mathbf{Z}^g}} [\ell_s (\mathbf{z}^g_1, \mathbf{z}^g_2)],
\end{equation}
where $\mathbf{z}^g_1 = f_{\theta_e}(\tilde{\mathbf{x}}^g_1)$, $\mathbf{z}^g_2 = f_{\theta_e}(\tilde{\mathbf{x}}^g_2)$, and $\mathbb{P}_{\mathbf{Z}^g}$ is the distribution of $\mathbf{z}^g$, where $\mathbf{z}^g \sim P(\mathbf{z}|\mathbf{x}^g)$.


\textbf{Local-to-global.} We use the global crops as ``anchor" points for their local crops because their larger crop size ensures that they will share some semantic content with the local crops. We, therefore, define a loss function that makes the local representations move closer to their corresponding global ones. Because, here, the global representations act as supervisory signals to the local ones, we either fix the global representations in the momentum encoder or stop their gradient in the back-propagation process. This yields the loss:
\begin{equation}\label{eqn:l2g}
    \mathcal{L}_{lg} =  \mathbb{E}_{\mathbb{P}_{\mathbf{Z}^g, \mathbf{Z}^l}} [\sum_{i=1,2} (\ell_s (\mathbf{z}^l_i, \text{sg}(\mathbf{z}^g_1))  + \ell_s (\mathbf{z}^l_i, \text{sg}(\mathbf{z}^g_2)))],
\end{equation}
where $\text{sg}(\cdot)$ stands for either the stop gradient operation in, e.g., SimSiam, or the momentum encoder in, e.g., MoCo.

\textbf{Local-to-local.} 
In the presence of complex image content, we expect two local views from the same image to often depict different semantic objects. Therefore, instead of encouraging local view similarity as in most existing works, we encourage their dissimilarity, thus preventing degenerate solutions where all local patches converge to the same representation independently of their content.
Given an affinity
function $\ell_{a}$, we express maximizing the local-to-local dissimilarity as minimizing the loss
\begin{equation}\label{eqn:l2l}
    \mathcal{L}_{ll} = \mathbb{E}_{\mathbb{P}_{\mathbf{Z}^l}} [\ell_{a} (\mathbf{z}^l_1, \mathbf{z}^l_2)].
\end{equation}
While one could in principle use any standard similarity measure, such as the cosine similarity, as an affinity function $\ell_a$, the high dimensionality of the feature space may lead to learning meaningless representations. Indeed, in high dimensional space, many directions allow one to push points away~\cite{aggarwal2001surprising}, and thus we need to find a direction that nonetheless encodes meaningful information.

To achieve this, we leverage the intuition that, although different images may contain local regions that depict the same semantic content, we expect on average local crops within an image to be more closely related than local crops from two different images. To encode this intuition, inspired by the Mutual Information Neural Estimator (MINE)~\cite{belghazi2018mine}, we make use of an auxiliary regressor
$f_{\theta_d}:\mathbb{R}^{n} \times \mathbb{R}^{n} \rightarrow \mathbb{R}^+$, which outputs a similarity value between two local crops. The parameters $\theta_d$ of this regressor are trained jointly with the other parameters of our approach. To this end, we seek to maximize the cost function
\begin{equation}\label{eqn:kernel}
  \Omega(\theta_d)  =\mathbb{E}_{\mathbb{P}_{\mathbf{Z}_1^l,\mathbf{Z}_2^l}} [f_{\theta_d}(\mathbf{z}^l_1, \mathbf{z}^l_2)] - \mathbb{E}_{\mathbb{P}_{\mathbf{Z}_1^l \otimes \mathbf{Z}^{l-}}}[ f_{\theta_d}(\mathbf{z}^l_1, \mathbf{z}^{l-})].  
\end{equation}

where $\mathbf{z}^{l-}$ is that of a local crop from a different image, which can be randomly sampled in the same batch and $\mathbb{P}_{\mathbf{Z}_1^l \otimes \mathbf{Z}^{l-}}$ is the product of two marginal distribution. By training it jointly with the encoder $f_{\theta_e}$, the regressor $f_{\theta_d}$ will adjust its similarity value based on the feature space distribution.

We then leverage the trained regressor to define our affinity function. That is, given
\begin{equation}
\theta_d^* = \argmax_{\theta_d}  \Omega(\theta_d)\;,
\end{equation}
we define
\begin{equation}
    \ell_a = f_{\theta_d^*}\;.
\end{equation}
We then use this definition of $\ell_a$ in the loss function of Eq.~\ref{eqn:l2l}. In other words, we train an affinity function to make local crops from the same image appear to be more similar than local crops from different images, and then train the encoder so as to minimize the resulting affinity between local crops from the same image to account for the fact that they will often depict different semantic contents.

Altogether, our SSL problem can therefore be formulated as the bi-level optimization problem
\begin{equation}
\begin{split}
  \min_{\theta_e}  \mathcal{L}_{gg}& +     \mathcal{L}_{lg} + \lambda \mathcal{L}_{ll}, \\
   \text{s.t.}  \quad \ell_a  & = f_{\theta_d^*} \\  \theta_d^* & = \argmax_{\theta_d}  \Omega(\theta_d),\label{eqn:cost_function}
\end{split}
\end{equation}
where $\lambda$ is a hyper-parameter balancing the dissimilarity and similarity terms, accounting for the fact that $\ell_a$ differs from the other terms. The details of our LoGo SSL strategy are provided in Algorithm~\ref{algorithm}.


\begin{algorithm}[tb]
  \caption{LoGo Pseudocode}\label{algorithm}
  \begin{algorithmic}
      \STATE {\bfseries Input:} batch size $N$, global and local augmentation $\tau^g$ and $\tau^l$, 
      \STATE {\bfseries Initialization:} encoder $f_{\theta_e}$, similarity measure $f_{\theta_d}$
      \WHILE{not reach epoch limits} 
      \STATE sample image minibatch  $\mathbf{X}$ 
      \FOR{j $<$ N}
      \STATE augment an image $\mathbf{X}(j)$ to get $\mathbf{x}_{1,2}^g$ and $\mathbf{x}_{1,2}^l$
      \STATE Get local and global representation, $\mathbf{z}_{1,2}^l(j)$ $\leftarrow$ $f_{\theta_e}(\mathbf{x}_{1,2}^l)$ and $\mathbf{z}_{1,2}^g(j)$ $\leftarrow$ $f_{\theta_e}(\mathbf{x}_{1,2}^g)$
      \ENDFOR
      \FOR{j $<$ N}
      \STATE Get positive local views pairs $\mathbf{z}_{1,2}^l(j)$
      \STATE Random pick a local view $\mathbf{z}_{1}^l(k), k \neq j$
      \STATE Maximize the loss by~\ref{eqn:kernel} and update the $f_{\theta_d}$
      
      \STATE Evaluate global views  similarity loss $\mathbf{z}_{1,2}^g(j)$ by~\ref{eqn:g2g}
      \STATE Evaluate local to global views similarity loss between $\mathbf{z}_{1,2}^g(j)$ and $\mathbf{z}_{1,2}^l(j)$ by~\ref{eqn:l2g}

      \STATE Evaluate the local to local loss for each $\mathbf{z}_{1,2}^l(j)$  as~\ref{eqn:l2l} 
      \STATE Minimize the total loss as~\ref{eqn:cost_function} and update the $f_{\theta_e}$
      \ENDFOR
      \ENDWHILE
      \STATE {\bfseries Output: The encoder network $f_{\theta_e}$} 
 \end{algorithmic}
\end{algorithm}

\section{Main Empirical Results}
We assess the performance and generality of our LoGo representation learning strategy by exploiting it within two popular SSL frameworks, namely MoCo~\cite{he2020momentum} and SimSiam~\cite{chen2021exploring}, and denote the resulting models as MoCo-LoGo and SimSiam-LoGo. We implemented our approach with Pytorch and run all the experiments on either 4 NVIDIA GeForce RTX 3090 or 2 NVIDIA V100 GPUs.

\subsection{Implementation Details}

\textbf{Optimization.} For our comparisons to be fair, we run all the SSL learning experiments for 200 epochs with a cosine learning decay scheduler~\cite{chen2020improved}, leading a learning rate $\eta_{t}=\eta_{\min }+\frac{1}{2}\left(\eta_{\max }-\eta_{\min } \right)\left(1+\cos \left( t\pi / T \right)\right) $. Following SimSiam and MoCo, we use the SGD optimizer and set the momentum value to 0.9 and the weight decay to $0.0001$. For MoCo, we find the best temperature values to be $\tau = 0.1, 0.2,0.07$ for CIFAR10, STL10 and IN-100, and keep the same learning rate and batch size as the original MoCo and SimSiam.

\textbf{Data augmentation.} We follow the same augmentation operations as in MoCo v2~\cite{chen2020improved}, including random cropping and resizing for the global and local views, random horizontal flipping, followed by random color jittering operations (brightness, contrast, saturation, and hue), and grayscale transformations. For ImageNet-100, we further add Gaussian blur.

\textbf{Regressor design.} The regressor $f_{\theta_d}$ consists of five $\text{fully-connected layer} + \text{synchronized batch normalization} + \text{ReLU}$ blocks, followed by a projection head and a soft-plus activation function that outputs a scalar value indicating similarity. We use the same structure for all datasets and experiments. Note that $\lambda$ in Eq.~\ref{eqn:cost_function} will be different for the different frameworks because they use different similarity losses. However, for each framework, we use the same $\lambda$ value for all datasets. Specifically, we fix $\lambda$ to $0.0005$ in MoCo-LoGo and $0.0001$ as in SimSiam-LoGo. In practice, the $\lambda$ is applied to the ratio of gradients between our regressor and backbone networks.

\subsection{Training and Evaluating the Features}

As a first set of experiments, we train the SSL models from scratch on CIFAR10~\cite{krizhevsky2009learning}, STL10~\cite{coates2011analysis}, and ImageNet100 (IN-100)~\cite{russakovsky2015imagenet,tian2020contrastive} with different backbone networks to show that our strategy is robust to image sizes and datasets scale. 

\textbf{Evaluation.} To evaluate the learned features on the respective validation sets, we use both a K-Nearest Neighbor (KNN) and a linear classifier. In the latter case, we train the linear classifier on the features extracted from the training set with the self-supervised pre-trained model. We train the classifier in the same way as in~\cite{he2020momentum,chen2020simple}. Details of datasets and parameter will be included in the supplementary.





As can be seen from Tables~\ref{tab:cifar},~\ref{tab:stl},~\ref{tab:IN}, our LoGo strategy consistently improves the classification accuracy of the baseline for both KNN and linear classification. In Table~\ref{tab:stl}, the performance of SimSiam drops significantly. This is because, unlike the average pooling in ResNets, the AlexNet used for this experiment relies on a fully connected layer outputting features in dimension 4096, which are difficult to handle by using cosine loss.
Interestingly, our SimSiam-LoGo nonetheless performs as well as MoCo-LoGo, which implies that our regressor provides valuable information to the encoder.

\begin{table}[]
\centering
\begin{tabular}{ccc}
    \toprule
             & KNN (acc \%)            & Linear (acc \%) \\ 
    \midrule
MoCo         & 79.58 & 80.37   \\ 
MoCo+LoGo    & \textbf{84.44} & \textbf{85.59}   \\ \midrule
SimSiam      & 80.48          &  83.02       \\ 
SimSiam+LoGo & \textbf{87.67} &   \textbf{88.02}  \\    
\bottomrule
\end{tabular}
\caption{Training on CIFAR10 with a ResNet-18 backbone. We show the top-1 accuracy for a KNN and a linear classifier.}\label{tab:cifar}
\end{table}

\begin{table}[]
\centering
\begin{tabular}{c|cc}
    \toprule
             & KNN (acc \%)            & Linear (acc \%) \\ \midrule
MoCo         & 72.13 &    79.07 \\    
MoCo+LoGo    & \textbf{76.79} &   \textbf{80.73} \\ \midrule
SimSiam      & 60.8          &    71.66     \\ 
SimSiam+LoGo & \textbf{76.96} &   \textbf{80.61}  \\    \bottomrule
\end{tabular}
\caption{Training on STL10 with a small Alexnet backbone. We show the top-1 accuracy for a KNN and a linear classifier.}\label{tab:stl}
\end{table}

\begin{table}[]
\centering
\begin{tabular}{ccc}
    \toprule
             & KNN (acc \%)            & Linear (acc \%) \\ \midrule
MoCo         & 64.18 & 68.48   \\ 
MoCo+LoGo    & \textbf{76.82} & \textbf{79.32}   \\ \midrule
SimSiam      & 71.21          &  75.48       \\ 
SimSiam+LoGo & \textbf{78.48} &  \textbf{80.94}    \\    \bottomrule
\end{tabular}
\caption{Training on ImageNet-100 with a ResNet-34 backbone. We show the top-1 accuracy for a KNN and a linear classifier.}\label{tab:IN}
\end{table}

\begin{table}[]
\centering
\begin{tabular}{lcc}
    \toprule
          & ResNet-34  & ResNet-50 \\ \midrule
          MoCo & 68.48 & 74.84 \\
MoCo-LoGo & 79.32 & 85.14     \\ \bottomrule
\end{tabular}
\caption{Linear classification accuracy (\%) of the Moco and MoCo-LoGo on IN-100 with ResNet-34 and ResNet-50 as backbone feature encoder.}\label{tab:compare_structure}
\end{table}

\begin{table*}[h]
\begin{tabular}{l|llllllllll}
    \toprule
           & CIFAR10        & CIFAR100      & Food & MIT67 & Pets  & Flowers & Caltech & Cars  & Aircraft & DTD   \\ \midrule
Super(IN-100) & \textbf{86.16} & 62.7          & 53.89   & 52.91 & 73.50 & 76.09   & 77.53      & 30.61 & 36.78    & 62.07 \\ \midrule
MoCo       & 83.71          & 60.59         & 58.21   & 57.54 & 64.30 & 85.56   & 74.12      & 32.63 & 46.23    & 60.64 \\
MoCo+Aug*  & 85.26          & \textbf{63.90} & 60.78   & 63.36 & 73.46 & 85.70   & 78.93      & 37.35 & 39.47    & 66.22 \\ \midrule
MoCo-LoGo w/o L2L  & 85.19          & 61.47         & 63.66   & 65.45 & 71.74 & 90.2   & 77.91      & 37.22 & 48.21    & 65.74 \\
MoCo-LoGo & 86.09 & 63.43 & \textbf{65.67} & \textbf{67.54} & \textbf{76.17} & \textbf{92.13} & \textbf{82.09} & \textbf{40.77} & \textbf{50.07} & \textbf{67.87} \\ \bottomrule
\end{tabular}
\caption{Transfer learning for image recognition. We report the recognition accuracy(\%). Super(IN-100) denotes the same network as for SSL but trained on IN-100 with supervision. MoCo+Aug*~\cite{lee2021improving} pre-trains the SSL encoder for 500 epochs, which is 300 epochs more than MoCo and our MoCo-LoGo.  MoCo-LoGo w/o L2L indicates our model without local-to-local dissimilarity. We highlight the best results in \textbf{bold}.}\label{tab:trans_linear}
\vspace{-0.2cm}
\end{table*}



\subsection{Transfer Learning}
One of the most important goals of representation learning is to obtain a backbone network extracting features that facilitate training on different datasets. We evaluate this on various datasets and downstream tasks. According to~\cite{chen2020simple,zhao2020makes}, MoCo constitutes the state-of-the-art for transfer to other datasets and tasks. In this context, most methods use a ResNet-50 as the backbone, and we thus train by applying our LoGo strategy to MoCo, i.e., MoCo-LoGo. Table~\ref{tab:compare_structure} shows that both MoCo and our MoCo-LoGo yield an improvement of around $6\%$ when increasing the capacity of the backbone from ResNet-34 to ResNet-50. Importantly, the advantage of MoCo-LoGo over MoCo remains unchanged compared to our previous experiments. 

To perform transfer learning, we, therefore, use our ResNet-50 pre-trained for 200 epochs. Below, we evidence the benefits of our approach for image recognition, object detection, and semantic segmentation using several datasets. For all the experiments in this section, we freeze the backbone networks and train the following task-dependent network modules according to the task at hand.

\subsubsection{Image Recognition}
Table~\ref{tab:trans_linear} compares the results of our approach and of the baselines on different image recognition datasets using the linear evaluation protocol of~\cite{lee2021improving,kornblith2019better,grill2020bootstrap}. Similarly to~\cite{lee2021improving}, we observed that the SSL strong augmentation methods and long training epochs harm the supervised baseline model for transfer learning. Therefore, we use the standard supervised training setting and train the model for 100 epochs. 
Since the image style and semantic classes of CIFAR10 and CIFAR100 are very similar to IN-100, the performance of MoCo-LoGo is very close to the supervised counterpart and to MoCo+Aug*, which was pre-trained for 500 epochs. Our method significantly outperforms the other baselines methods, especially on fine-grained classification datasets, such as Flowers, Aircraft, Caltech, and Food. More information about the datasets can be found in the supplementary material. Altogether, these results show that our MoCo-LoGo enables the backbone to capture rich semantic information.

\subsubsection{Dense Prediction}
We use our SSL trained networks to conduct object detection experiments on both MS-COCO~\cite{lin2014microsoft} and PASCAL VOC~\cite{everingham2010pascal},  as well as semantic segmentation experiments on MS-COCO~\cite{lin2014microsoft}. We employ the Average precision (AP) to evaluate the results. Specifically, we use $\text{AP}_{50}$ and $\text{AP}$, $\text{AP}_{75}$, where 50 and 75 indicate the that IoU threshold is set to 0.5 and 0.75, respectively, and $\text{AP}$ denotes the average precision when the IoU threshold is varied from 0.5 to 1.0.

\textbf{Setting}. For these experiments, we follow one of the protocols in~\cite{he2020momentum}. We adopt the pretrained ResNet 50-C4 as the backbone and fine-tune the Faster R-CNN~\cite{ren2015faster} detector on both the VOC and COCO datasets. We apply $2 \times$ schedulers on both datasets, which means that we train for approximately 23 epochs. In the VOC dataset, we use the 07+12 training sets to train the detector and the VOC \textit{test2007} as test set; for COCO, we train on the \textit{train2017} set (around 118k images) and evaluate on \textit{val2017}. 

As shown in the left portion of Table~\ref{tab:trans_dense}, our MoCo-LoGo achieves the best results in terms of $\text{AP}$ and $\text{AP}_{75}$ for the detection task. Note that these metrics are more strict than $\text{AP}_{50}$. On the segmentation task, our method achieves the best results in all three metrics. 
Remarkably, MoCo-LoGo also surpasses the supervised backbones trained on IN-1k in terms of the more strict metrics $\text{AP}$ and $\text{AP}_{75}$. This indicates that a small number of images suffice for our strategy to capture important semantic information.

\begin{table*}[h]
\centering
\begin{tabular}{l|ccc|ccc|ccc}
    \toprule
                & \multicolumn{6}{c|}{Object Detection}                                                              & \multicolumn{3}{c}{Segmentation}                      \\ \cline{2-10}
                & \multicolumn{3}{c|}{VOC07+12}                    & \multicolumn{3}{c|}{MS-COCO}                        & \multicolumn{3}{c}{MS-COCO}                         \\ \cline{2-10}
                & $\text{AP}^{bb}_{50}$          & $\text{AP}^{bb}$             & $\text{AP}^{bb}_{75}$          & $\text{AP}^{bb}_{50}$          & $\text{AP}^{bb}$             & $\text{AP}^{bb}_{75}$           & $\text{AP}^{mk}_{50}$           & $\text{AP}^{mk}$             & $\text{AP}^{mk}_{75}$        \\ \hline
Super(IN-1k) & \textbf{81.30} & 53.50           & 58.80           & \textbf{59.90} & 40.00             & 43.10           & 56.50           & 34.70           & 36.90           \\
MoCo            & 78.65         & 52.43          & 57.22          & 59.11         & 39.38          & 42.55          & 55.83          & 34.52          & 36.84          \\ 
MoCo-LoGo       & 81.12         & \textbf{54.91} & \textbf{61.06} & 59.74         & \textbf{40.23} & \textbf{43.48} & \textbf{56.55} & \textbf{35.04} & \textbf{37.42} \\ \bottomrule

\end{tabular}
\caption{Transfer learning on object detection and semantic segmentation. Super (IN-1k) indicates supervised training on ImageNet-1k. MoCo and MoCo-LoGo are trained on IN-100. The best entries are shown in \textbf{bold}.}
\vspace{-1em}
\label{tab:trans_dense}
\end{table*}

\section{Ablation Study}
In this section, we study the components introduced in our strategy to validate their functionality. Since stop-gradient, batch size and learning rate have been intensively studied by~\cite{he2020momentum,chen2020simple}, we will not discuss them here. We focus our analysis on our main contributions and on what the regressor learns.

\subsection{Muti-crop and Similarity Loss}
First, we remove the local-to-local dissimilarity term $\ell_a$ on the local patches to study the improvements contributed by our multiple crops and similarity loss used to encode global-to-global and local-to-global relationships. We use w/o L2L to denote this baseline. 
 
We report the Top-1 KNN accuracy of w/o L2L and our full strategy in Table~\ref{tab:ablation}. Note that only performing multi-crop with similarity loss in the pretaining stage yields a better KNN accuracy than vanilla MoCo and SimSiam. However, there is still a gap with our full model. Consistent phenomena can be seen in Table~\ref{tab:trans_linear}, where we transfer the MoCo-LoGo w/o L2L pretrained model to other image recognition tasks. In some datasets, such as Pets and Caltech, it still falls behind our full strategy by a margin. This evidences the importance of encouraging dissimilarity between the local crops for fine-grained classification tasks.
 
\subsection{Learnable Affinity Measure}
To analyze what our regressor learns, we study the difference between employing our regressor or the cosine distance. We observed that maximizing the cosine distance between the local crops during training causes the model to collapse. In other words, when we use KNN to monitor the progress of training, the accuracy rapidly drops. 
To further analyze our learned measure, we compare the similarity of different images computed by using either the cosine distance or our regressor. 

\begin{figure}[!h]
\begin{subfigure}{1.0\linewidth}
\includegraphics[width=1.0\linewidth]{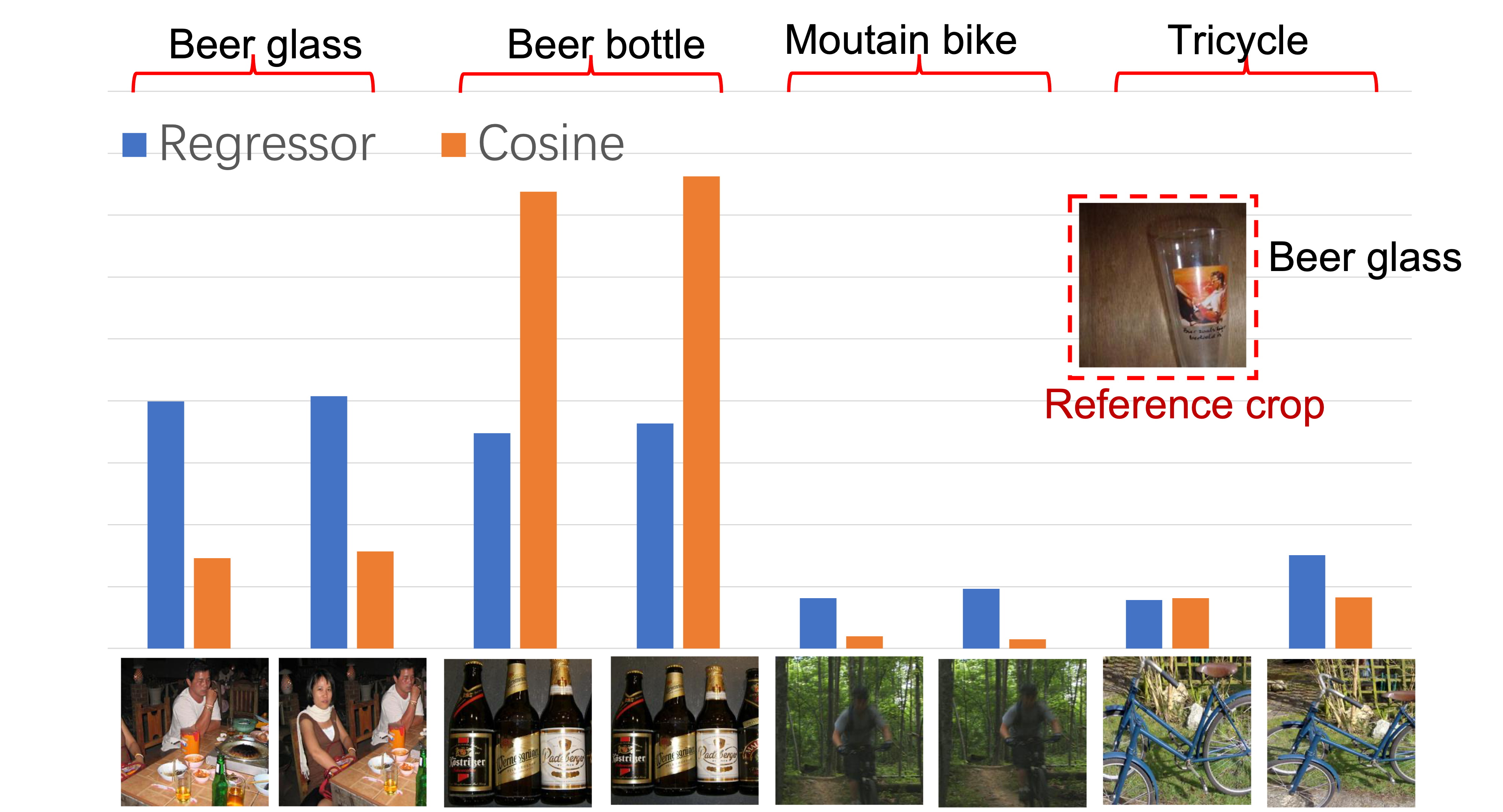}
\end{subfigure}
\begin{subfigure}{1.0\linewidth}
\includegraphics[width=1.0\linewidth]{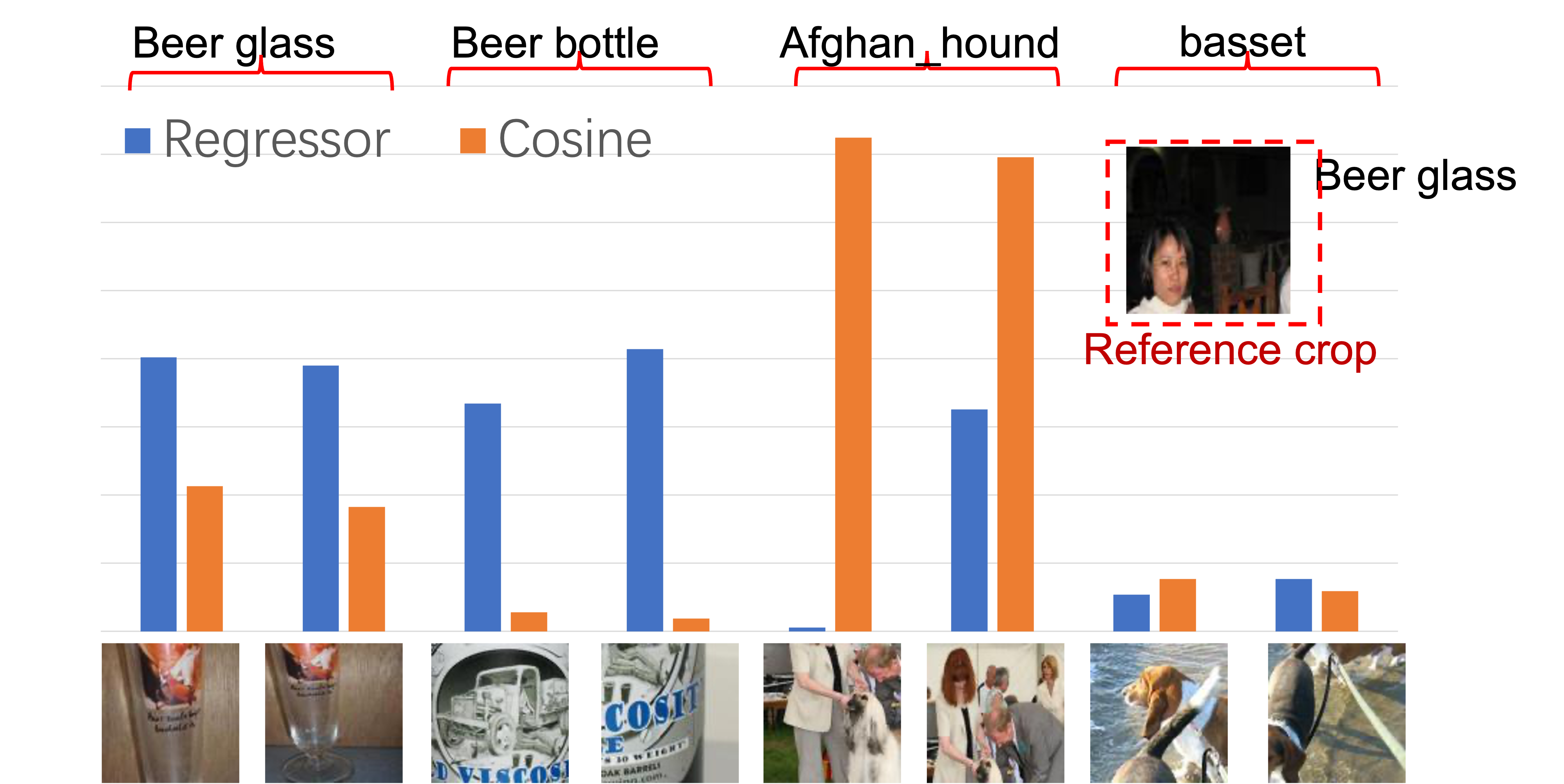}
\end{subfigure}
\caption{Comparison of our learned affinity measure and the cosine similarity. The values are the normalized regressor and cosine similarity between the reference crops and every crop on the $x$-axis. Higher values indicate a larger similarity. }\label{fig:discriminator}
\vspace{-0.3cm}
\end{figure}

To this end, we randomly obtain a crop from an image and compute both the cosine and regressor similarity with other 40 different crops (10 crops from 4 different images).
In Figure~\ref{fig:discriminator}, we visualize such similarities for 2 crops per class. As shown in the top portion of the figure, taking the beer glass crop as a reference, the cosine similarity wrongly assigns a larger similarity to the beer bottle than to its own class. By contrast, our regressor correctly preserves the beer glass information in such a complex scene, where the glass is not as salient as the people. In the lower portion of the figure, we show the similarities for a reference crop that depicts two different objects. In this case, the cosine distance fails and focuses on the person only. On the contrary, our learned affinity measure gives a higher similarity to the reference class and the semantically-close class. Interestingly, it yields very different values for the two different crops belonging to the Afghan hound class; a higher value for the crop where the person's face is visible, matching the fact that the reference image also contains a human face. More results are provided in the supplementary material. They further support the evidence that our learned measure encodes valuable semantic similarities between crops, thus providing stable supervision for the encoder.
 

\begin{table}
\begin{tabular}{llccc}
    \toprule
                         &      & CIFAR10        & STL10          & IN100          \\ \midrule
\multirow{2}{*}{MoCo-LoGo}    & w/o L2L & 82.31          & 74.26          & 69.00             \\
                         & full  & \textbf{84.44} & \textbf{76.79} & \textbf{76.82} \\  \midrule
\multirow{2}{*}{SimSiam-LoGo} & w/o L2L & 83.33          & 60.21          & 76.64          \\
                         & full & \textbf{87.67} & \textbf{76.96} & \textbf{78.48} \\ \bottomrule
\end{tabular}
\caption{Comparison of the top-1 KNN classification accuracy (\%) for the full LoGo strategy and LoGo without local-to-local strategy on both MoCo and SimSiam. The models are trained and tested on the same dataset.}\label{tab:ablation}
\vspace{-0.2cm}
\end{table}

\section{Conclusion and Limitation}

We have presented a new SSL strategy that leverages local and global views so as to better account for complex visual content. Our approach generalizes to existing SSL frameworks, consistently boosting their performance. Our learning strategy not only enables the global crops to preserve the invariant semantic information but allows the local crops to have diverse representations, thus not destroying their semantic meaning.  Our extensive experiments have demonstrated the effectiveness of our strategy, further confirming the importance of every component in our approach. Our learnable affinity measure incurs more computation and parameters, however, it captures semantically-driven similarities between image patches and thus has the potential to be applied to other downstream tasks. 

\section{Acknowledgements}
This work was supported in part by the Swiss National Science Foundation via the Sinergia grant CRSII5$-$180359. C Qiu and W Ke were supported by National Natural Science Foundation of China under Grant No. 62006182.






{\small
\bibliographystyle{ieee_fullname}
\bibliography{egbib}
}

\end{document}